# Fast Object Classification in Single-pixel Imaging

Shuming Jiao

Tsinghua Berkeley Shenzhen Institute (TBSI), Shenzhen, 518000, China

albertjiaoee@126.com

**ABSTRACT**

In single-pixel imaging (SPI), the target object is illuminated with varying patterns sequentially and an intensity sequence is recorded by a single-pixel detector without spatial resolution. A high quality object image can only be computationally reconstructed after a large number of illuminations, with disadvantages of long imaging time and high cost. Conventionally, object classification is performed after a reconstructed object image with good fidelity is available. In this paper, we propose to classify the target object with a small number of illuminations in a fast manner for Fourier SPI. A naive Bayes classifier is employed to classify the target objects based on the single-pixel intensity sequence without any image reconstruction and each sequence element is regarded as an object feature in the classifier. Simulation results demonstrate our proposed scheme can classify the number digit object images with high accuracy (e.g. 80% accuracy using only 13 illuminations, at a sampling ratio of 0.3%).

## 1. INTRODUCTION

Single-pixel imaging [1, 2] (SPI) is a computational imaging technique in which the object scene is illuminated with varying patterns sequentially and a one-dimensional intensity sequence is recorded by a single-pixel detector without spatial resolution. The object image is computationally reconstructed from the single-pixel intensity sequence and the illumination patterns. Single-pixel imaging has several unique features compared with conventional imaging techniques, such as low-cost single-pixel detector instead of pixelated detector, imaging without direct line of sight, and imaging under weak light condition. Single-pixel imaging is proposed for a variety of applications such as terahertz imaging [3], remote sensing [4], three-dimensional (3-D) imaging [5-7], microscopy [8, 9], scattering imaging [10], hyperspectral imaging [11], X-ray imaging [12], optical security [13-16], lidar detection [17, 18] and gas leak monitoring [19].

On the downside, SPI usually requires a large number of illuminations with different patterns to acquire a reconstructed object image with satisfactory quality. For example, in computational ghost imaging, where random intensity patterns are employed as the illumination patterns and correlation is used as the reconstruction algorithm, it may require more than 10000 illumination patterns to capture an object image with a size of 64×64 pixels. Some attempts have been made to improve the imaging efficiency and reduce sampling ratio in SPI by using more advanced reconstruction algorithms (e.g. compressive sensing [20, 21], deep learning [22-25]) or designing more appropriate illumination patterns (e.g. Fourier transform patterns [26], Hadamard patterns [8], wavelet patterns [27]). In addition, error correction coding ([28-30]) can be used to improve the imaging efficiency in encrypted SPI. However, it is still hard to successfully recover an object image with good fidelity from only a short single-pixel intensity sequence consisting of a small number of pixels.

In various optical imaging systems including SPI, the automatic classification of target objects is a critical issue and has wide applications. For example, when SPI is applied in microcopy applications (such as medical diagnosis and water quality inspection), it is necessary to identify what category of microorganisms or cells the sample belongs to. When SPI is applied in remote sensing, the ground objects (buildings, grassland, forest etc.) need to be automatically classified. Conventionally, the target object classification can be performed after a high-quality object image is already computationally reconstructed from a SPI system. Then the numerous image classification algorithms in computer vision field (such as [31-33]) can be directly employed and there is no difference in the image classification process between photographs captured by a camera and reconstructed images from SPI. However, as stated above, a high-quality reconstructed image can only be obtained after a large number of illuminations with high cost. In this paper, we investigate the fast classification of target object in SPI when only a very limited number of illuminations is performed and a reconstructed image with reasonable quality is not available.

In some previous works, the authentication (or recognition) of target object was implemented for SPI with relatively less number of illuminations [14-16, 34-36]. However, despite the success in these works, the target object can only be authenticated as correct or wrong, regarding to a reference object image (or possibly some geometrical variant of the reference object [37]). A classification of multiple-class objects cannot be achieved in these schemes [14-16, 34-36]. Furthermore, the number of illuminations for successful authentication is still at least 3% of the Nyquist limit [15], and is yet to be improved.

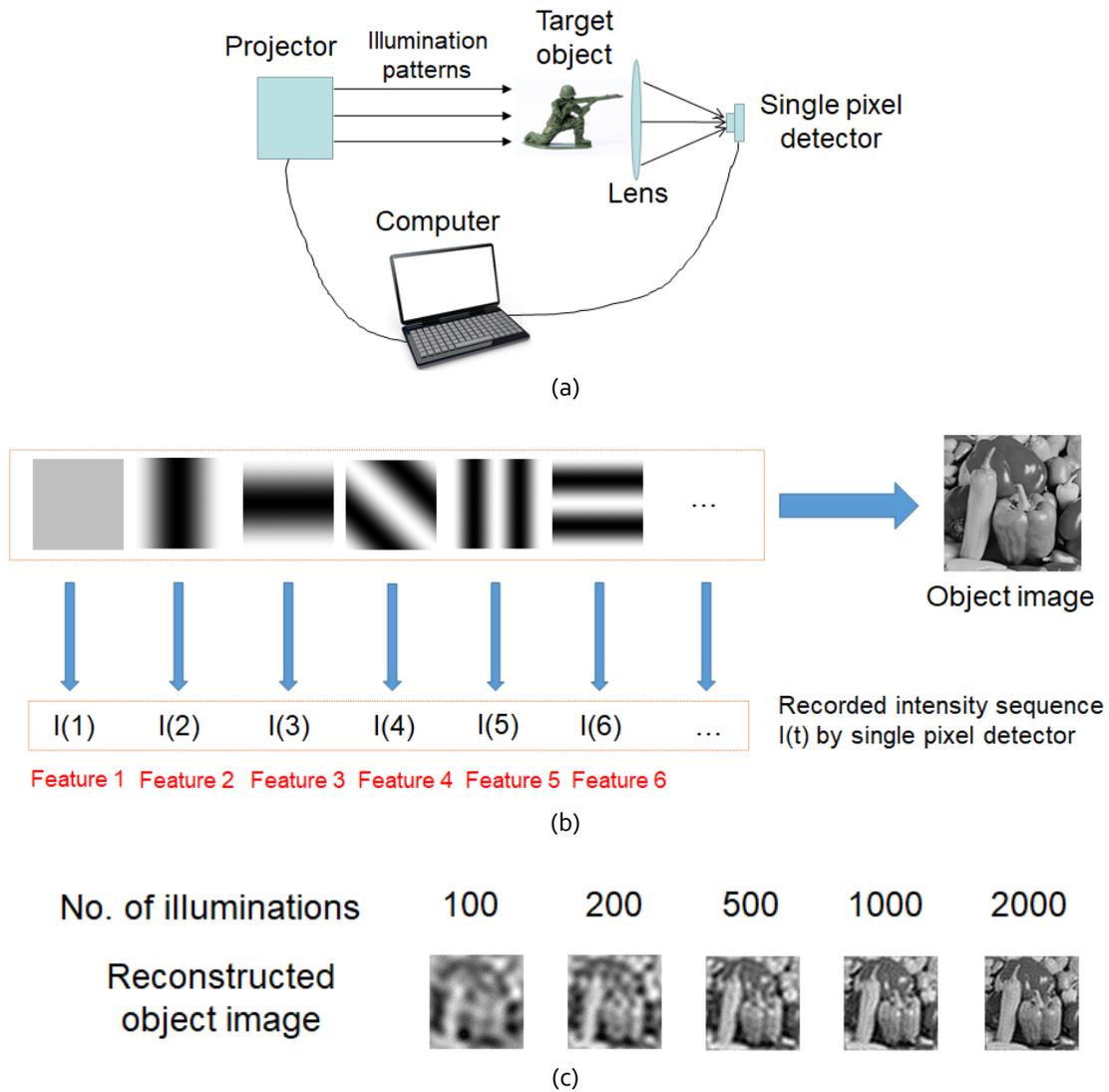

Figure 1. (a) Optical setup for a Fourier single-pixel imaging (FSPI) system; (b) Working mechanism of a Fourier single-pixel imaging system; (c) Reconstructed image quality grdually enhanced as the number of illumiantions increases

## 2. PROPOSED FAST OBJECT CLASSIFICAION SCHEME

In this work, the author proposes to employ a naive Bayes classifier [38] to perform target object classification in Fourier single-pixel imaging (FSPI) [26] with a very small number of illuminations and without object image reconstruction. The optical setup of a FSPI system is shown in Figure 1(a). A comparison of our proposed scheme with conventional object classification scheme in SPI is demonstrated in Figure 2.

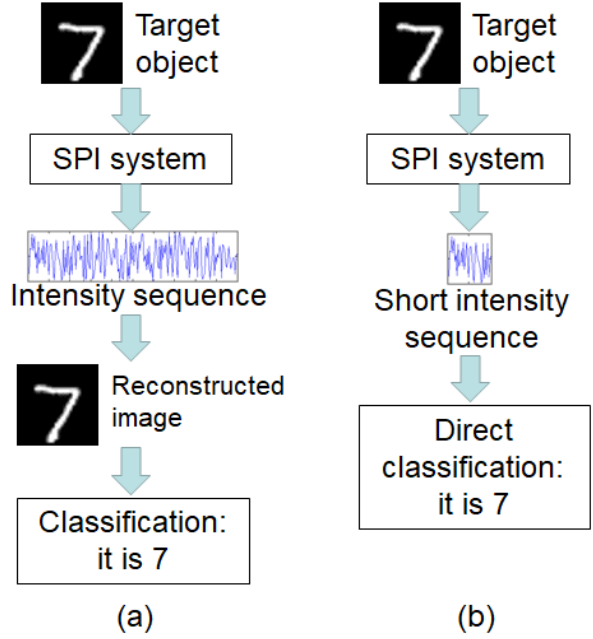

Figure 2. (a)Conventional approach for target object classification in single-pixel imaging; (b) Proposed approach for target object classification in single-pixel imaging in our work.

In FSPI, the sinusoidal patterns in Fourier transform is projected onto the target object sequentially from low frequency components to high frequency components. Then the total light intensity I(t) for the object scene under each illumination is recorded by a single-pixel detector, which is equivalent to the inner product between the object image $O(x, y)$ and the sinusoidal patterns $\cos[-2\pi(ux + vy)]$ or $\sin[-2\pi(ux + vy)]$ [(x,y): spatial domain; (u,v): frequency domain], illustrated by Equation (1)-(3) and Figure 1(b), where $\text{Re}[F(u, v)]$ and $\text{Im}[F(u, v)]$ denote the real part and imaginary part of a Fourier spectrum coefficient. In this way, the Fourier spectrum coefficients of the target object is gradually captured point by point in SPI.

$$F(u, v) = \iint O(x, y) \exp[-2\pi(ux + vy)] \, dxdy \tag{1}$$
$$\text{Re}[F(u, v)] = \iint O(x, y) \cos[-2\pi(ux + vy)] \, dxdy \tag{2}$$
$$\text{Im}[F(u, v)] = \iint O(x, y) \sin[-2\pi(ux + vy)] \, dxdy \tag{3}$$

For an object image of size M×N, the number of components in its entire discrete Fourier spectrum will be M×N complex values. Due to the symmetric property in Fourier spectrum, only half of the spectrum needs to be recorded. One complex value can be acquired from two recorded real intensity values using two illuminations. So it requires M×N illuminations (full sampling) to acquire the entire Fourier spectrum of the object image. Since most natural object images are locally smooth, the energy of their Fourier spectrum is concentrated in the low frequency components. The reconstructed object image quality can be gradually enhanced as the number of illuminations increases. An object image with reasonable quality can usually be reconstructed with a sampling ratio of around 10%-20% (10%-20% lowest frequency components in the spectrum). In this work, the target object classification is performed when the sampling ratio is much smaller than 10%.

A Naive Bayes probabilistic classifier is a classical data classification algorithm in machine learning and statistics. In single-pixel imaging, it is assumed that the detector records a sequence of N intensity values I(t) ($1 \leq t \leq N$), corresponding to the real parts and imaginary parts of each complex spectrum components. Each recorded single-pixel intensity value I(t) can be considered as one feature (or predictor) of the target object $O(x, y)$. The target object is classified into one of the K possible categories based on the N features. Naive Bayes classifier is a supervised machine learning approach and a set of exemplar object images with correct class labels are required as training images. First, from the FSPI results of training objects, the intensity values of each feature (such as $\text{Re}[F(1,1)]$ or $\text{Im}[F(2,3)]$) will have K different probability distribution models for K different categories of objects. A Gaussian probability distribution (Equation

(4)) with parameters $\mu$ (average intensity) and $\sigma$ (standard deviation of intensities) can be constructed with data fitting for each intensity sequence element I(t) ($1 \le t \le N$) and each object class C(m) ($1 \le m \le K$), from the training images. There are totally N × K Gaussian probability distribution functions.

$$P(I(t)|C(m)) = \frac{1}{\sqrt{2\pi}\sigma_{t,m}} \exp\{-\frac{[I(t)-\mu_{t,m}]^2}{2\sigma_{t,m}^2}\} \quad (4)$$

For an arbitrary given intensity sequence $I$ (I(1), I(2),…, I(N)) generated from a target object (not belonging to the training set) and recorded by a FSPI system, the probability of $I$ belonging to each object category can be calculated by combining the probabilities of each element I(t) in the sequence belonging to each object category. The overall probability of the entire sequence belonging to one object category $C(m)$ can be calculated by Equation (5)

$$P(C(m)|I) = P[C(m)] \prod_{t=1}^{N} P(I(t)|C(m)) \quad (5)$$

where $P[C(m)]$ is the proportion of $C(m)$ category samples in the training data set and $P(I(t)|C(m))$ is the probability of the sequence element $I(t)$ belong to $C(m)$ (obtained from the pre-calculated probability distribution functions). Finally, the classification result is determined by selecting the maximum one among the m probability values of $P[C(m)]$. For example, if $P[C(2)]$ has the highest value, the intensity sequence is considered being from an object of second category.

## 3. SIMULATION RESULTS

The MNIST database [39], consisting of huge numbers of handwritten number digit images (Figure 3), is employed to verify our proposed scheme. Each number digit image is considered as one possible target object in FSPI and they need to be classified into ten categories (digits 0 to 9). In simulation, 9000 images are employed as training objects and 500 images are employed as the testing objects. All training and testing images are scaled to a size of $64 \times 64$ pixels.

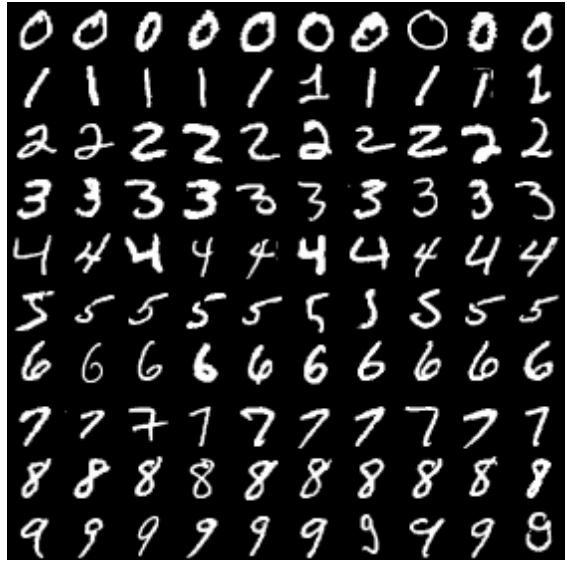

Figure 3. Some image examples [40] of MINST database

In our simulation, a single-pixel intensity sequence is recorded by a FSPI system for each training and testing object. A naive Bayes classifier system is trained with the intensity sequence data generated from labeled training objects. Then the intensity sequence data generated from testing objects are classified with the classifier. The overall classification accuracy (correct rate) versus the number of illuminations in FSPI for testing objects is shown in Figure 4.

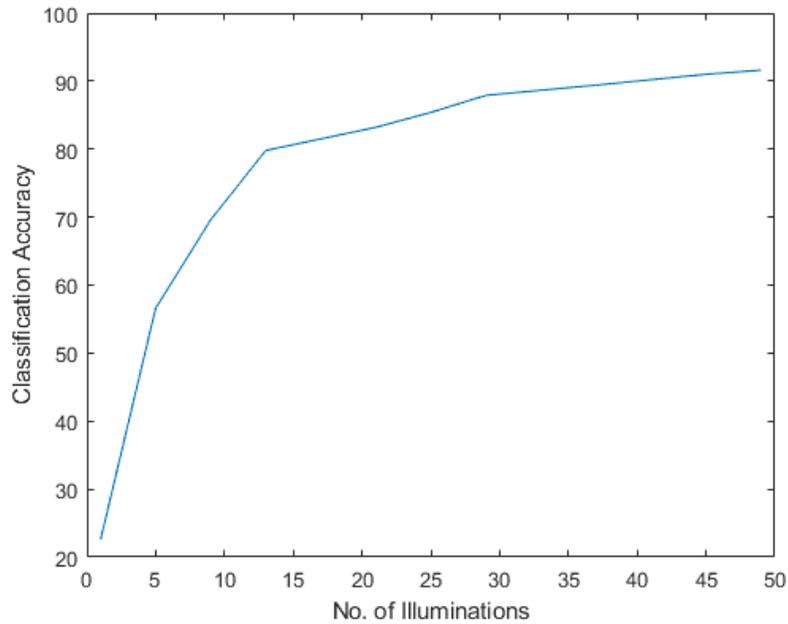

Figure 4. Classification accuray versus number of illumiantion patterns in FSPI for number digit obejcts with an image size of 64 × 64 pixels

In FSPI, it usually requires approximately 4096 (64 × 64) illuminations to record the entire Fourier spectrum at full sampling ratio for an object image with a size of 64 × 64 pixels. A minimum of 100 to 200 illuminations is usually required to ensure a reconstructed image with reasonable quality. However, our proposed scheme can achieve a classification accuracy of over 80% with only 13 illuminations (a sampling ratio of only 0.3%) and a classification accuracy of nearly 90% with around 50 illuminations. Fifteen examples of original testing object images, classification results and reconstructed image results with inverse Fourier transform (when the number of illuminations is 13) are shown in Figure 5.

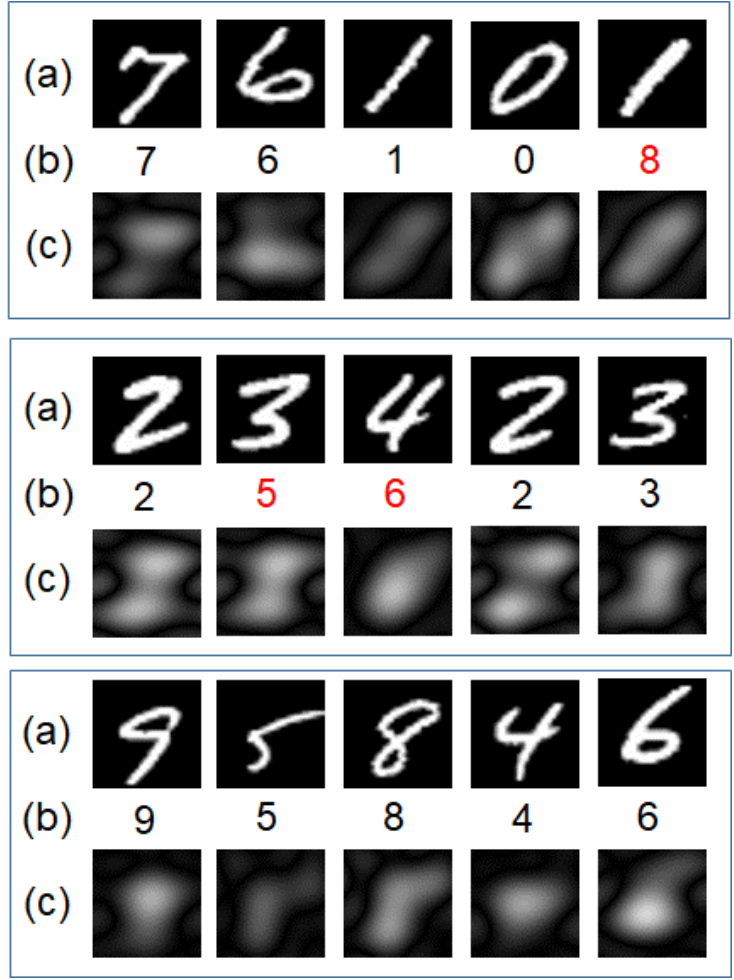

Figure 5. Fifteen examples of (a) original testing object images; (b) classification results (correct ones marked with black color and wrong ones marked with red color); and (c) reconstructed image results with inverse Fourier transform (when the number of illuminations is 13)

It can be observed that the quality of reconstructed images (shown in Figure (c)) are very low and the number digit is very difficult to be identified visually when the number of illuminations is only 13. However, our proposed scheme can yield correct classification results for 80% of the objects only based on the 13 elements in the single-pixel intensity sequence, without any image reconstruction.

## 4. CONCLUSION

In single-pixel imaging (SPI), a high quality object image can only be reconstructed after an adequate number of illuminations, with disadvantages of long imaging time and high cost. We proposed to classify the target object with a small number of illuminations in a fast manner for SPI. A naive Bayes classifier is employed to classify the target objects based on the single-pixel intensity sequence acquired by a Fourier single-pixel imaging (FSPI) system and each intensity is regarded as an object feature in the classifier. Simulation results demonstrate our proposed method can classify the number digit objects images with high accuracy (e.g. 80% accuracy can be achieved using only 13 illuminations). Our work reveals the possibility of performing high-level computer vision tasks (e.g. image classification) even when a good reconstructed result is not available in a SPI system.

The results and discussions in this paper can provide a basis that numerous future works are built upon. For example, more advanced classification algorithms (e.g. deep learning), in addition to Bayes classifier, can be attempted for the object classification problem in SPI. Results from real optical experiments, in addition to numerical simulation, can be evaluated.

The classification of more complicated objects in practical applications (such as classification of cells in single-pixel microscopy and classification of land objects in single-pixel remote sensing), in addition to simple number digit objects, can be furthered investigated.